\documentclass[sigconf, table, dvipsnames]{acmart}

\usepackage{listings}
\usepackage{booktabs}
\usepackage{array}
\usepackage{multirow}
\usepackage{enumitem}
\usepackage{lineno}
\usepackage{subcaption}
\usepackage{tikz}
\usepackage{tabularx}
\usepackage{placeins} 

\usepackage{algorithm}
\usepackage{algpseudocode}

\usetikzlibrary{
  positioning,
  arrows.meta,
  shapes.geometric,
  fit,
  backgrounds,
  calc
}

\DeclareUnicodeCharacter{2264}{\ensuremath{\leq}}

\renewcommand\footnotetextcopyrightpermission[1]{} 

\AtBeginDocument{%
  \providecommand\BibTeX{{%
    \normalfont B\kern-0.5em{\scshape i\kern-0.25em b}\kern-0.8em\TeX}}}

\graphicspath{{./images/}} 
\begin{document}

\title{SCM-based Fairness and Faithful Explainability for Legal Document Classification}

\author{Yasmina El Kacemi}
\affiliation{%
  \institution{University of Amsterdam}
  \city{Amsterdam}
  \country{The Netherlands}
}
\email{yasmina.el.kacemi@student.uva.nl}

\author{Seyed Sahand Mohammadi Ziabari}
\authornote{Corresponding author.}
\affiliation{%
  \institution{University of Amsterdam}
  \city{Amsterdam}
  \country{The Netherlands}
}
\email{s.s.mohammadiziabari@uva.nl}

\author{Ali Mohammed Mansoor Alsahag}
\affiliation{%
  \institution{University of Amsterdam}
  \city{Amsterdam}
  \country{The Netherlands}
}
\email{a.m.m.alsahag@uva.nl}

\begin{abstract}

{
Transformer models such as LegalBERT are increasingly used in legal decision support. This raises paired concerns about fairness and the transparency of model explanations, which are usually evaluated separately. It therefore remains open whether a debiasing intervention that changes fairness also changes how faithfully explanations reflect the model's reasoning. This paper investigates that question on the ECtHR alleged-violations corpus from LexGLUE. It compares a LegalBERT baseline against a fairness-regularised variant that penalises stereotypical warmth and competence representations during fine-tuning. The comparison covers predictive performance, demographic fairness, and SHAP explanation faithfulness, over five random seeds. At the performance-optimal strength, the intervention does not reduce demographic disparity. This null holds across two fairness definitions and a conventional word-pair control on the gender axis where that paradigm is designed to apply. Classification performance is unchanged. However, the intervention degrades explanation sufficiency across all five seeds and three thresholds, with non-overlapping bands. A shuffled-pair control reproduces this degradation while leaving performance and fairness unchanged, so the effect follows from contrastive regularisation of any kind, not the warmth and competence structure. The contribution is a dissociation: a change in explanation behaviour did not signal a change in fairness, so fairness must be measured directly.
}

\end{abstract}



\fancyhead{}
\maketitle
\section{Introduction}
\label{sec:introduction}

The application of transformer-based language models to legal natural
language processing (legal NLP) has substantially expanded the capacity
for automated analysis of court decisions and statutory
texts~\cite{chalkidis2022}. Domain-adapted models such as LegalBERT,
pretrained on large legal corpora, achieve strong performance on legal
document classification, a central benchmark task for evaluating legal
AI systems~\cite{chalkidis2020legalbert}. As these systems move from
research settings into decision-support roles, two properties become
critical beyond predictive accuracy: fairness, the absence of systematic
discrimination across demographic groups, and transparency, the degree
to which model predictions can be accurately explained.

Both pose unresolved challenges in legal NLP. Models trained on
historical legal corpora absorb the institutional and social biases
embedded in those texts and reproduce them in downstream
predictions~\cite{caliskan2017semantics, gallegos2024survey}. Such bias
need not arise from explicit demographic attributes: the Stereotype
Content Model (SCM) holds that group perception is organised around two
latent dimensions, warmth and competence, which are geometrically encoded
in language model representations and influence behaviour even when
protected attributes are absent~\cite{omrani2023scm}. Standard debiasing
targeting specific word pairs is therefore insufficient, addressing
surface proxies while the underlying structure persists~\cite{gonen2019lipstick};
SCM-based debiasing offers a more principled, group-agnostic alternative
for the broad and uneven range of groups in legal documents~\cite{ungless2022robust}.

Transparency is equally fraught. Post-hoc methods such as
SHAP~\cite{lundberg2017unified} and LIME~\cite{ribeiro2016lime} are widely
used, yet a plausible-looking explanation need not be a faithful one:
Jacovi and Goldberg~\cite{jacovi2020towards} distinguish plausibility from
faithfulness, the degree to which an explanation reflects the model's
actual reasoning. In legal contexts an unfaithful explanation can conceal
biased or arbitrary reasoning while projecting principled
transparency~\cite{atanasova2023faithfulnesstestsnaturallanguage}. Since ECtHR judgments bear
directly on individual rights, fairness and faithful explainability are
conditions for legal accountability, not only technical requirements.

The relationship between these two properties remains poorly understood.
The debiasing literature evaluates fairness in isolation, while
explainability research focuses on interpretability or plausibility
without asking whether debiasing affects how reliably explanations
reflect model reasoning. Whether the two can be improved together, or are
independent outcomes of the same intervention, is an open empirical
question with direct implications for how high-stakes AI systems should
be evaluated. The closest work is Brandl et al.~\cite{brandl2024interplay},
who study fairness and explainability on the same ECtHR corpus and find
that fine-tuning-based bias mitigation does not consistently produce
fairer models and that fairness and explainability are largely
orthogonal; their study evaluates rationale plausibility through human–model alignment and sufficiency through AOPC. Whether SCM-based debiasing affects how accurately
post-hoc explanations reflect model reasoning has not been examined.

This paper investigates, through a controlled pre-specified study,
whether SCM-based fairness regularisation applied to LegalBERT affects
(a) demographic disparity in predictions, (b) the faithfulness of SHAP
explanations, and (c) classification performance, on the ECtHR corpus
from LexGLUE~\cite{chalkidis2022}. Throughout, transparency is understood
as explanation faithfulness: the degree to which a post-hoc explanation
reflects the model's actual reasoning rather than producing a plausible
account of it~\cite{jacovi2020towards}.

\medskip
\noindent
The contributions of this paper are as follows:
\begin{itemize}
  \item A multi-seed evaluation showing SCM-based fairness regularisation
        does not reduce demographic disparity on ECtHR, a null holding
        across two fairness definitions, on the more reliable ethnicity
        axis, and under an independent named-entity grouping
        (Section~\ref{sec:results}).
  \item Evidence that the same intervention consistently degrades SHAP
        explanation sufficiency across five seeds, at no reliable cost to
        performance (Section~\ref{sec:results}).
  \item A shuffled-pair control showing none of these effects is specific
        to the warmth--competence structure; an arbitrary contrastive
        penalty reproduces them (Section~\ref{sec:results}).
  \item The resulting dissociation: fairness and explanation faithfulness
        move independently, so explanation quality cannot serve as a proxy
        for fairness in legal-NLP auditing in this setting
        (Section~\ref{sec:discussion}).
\end{itemize}

This study investigates whether SCM-based fairness regularisation can alter
demographic disparities and the faithfulness of post-hoc explanations in
legal document classification without materially compromising predictive
performance. Using a standard LegalBERT classifier as the baseline, we
evaluate the effect of the proposed regularisation on three complementary
dimensions: predictive performance, demographic fairness, and explanation
faithfulness. Fairness is assessed using Demographic Parity Difference
(DPD) and Equalized Odds Difference (EOD), with Disparate Impact (DI)
reported as a diagnostic measure, while explanation faithfulness is
evaluated through sufficiency and comprehensiveness.

In addition to comparing the regularised model with the unregularised
baseline, we examine how these outcomes vary across different
regularisation strengths $\lambda$. This analysis is intended to determine
whether stronger representational constraints produce systematic changes in
fairness, predictive performance, or explanation behaviour, and to identify
whether improvements in one dimension are accompanied by gains or
trade-offs in the others.

\section{Related Work}
\label{sec:related_work}
 
 

Fairness and explainability have received substantial attention in NLP, but their interaction remains underexplored. This section reviews the limited work that treats them jointly, then outlines the three technical foundations of this study: the Stereotype Content Model for group-agnostic debiasing, faithfulness as an explainability criterion for high-stakes auditing, and fairness metrics for measuring disparity. Together, these strands define the research gap addressed in this paper.
\subsection{Fairness and Explainability as Joint Concerns}
\label{sec:rw-joint}

Although both fairness and explainability are well studied in NLP, research treating them as jointly examined objectives is sparse. The dominant paradigm evaluates them independently: fairness with group metrics such as Demographic Parity Difference and Disparate Impact, explanation methods on plausibility or faithfulness, with little attention to how one affects the other. \citet{balkir2022challenges} survey explainability methods applied in fairness contexts and find that how explainability actually mitigates bias is rarely specified, arguing for more principled joint treatments.

The most closely related empirical work is \citet{brandl2024interplay}, who conduct a first study of the interplay between fairness and explainability on two English datasets: BIOS (occupation classification, gender as the protected attribute) and ECtHR (violation prediction, defendant state binarised as Eastern European versus the Rest). Using five fairness-promoting and three rationale-extraction methods on hierarchical RoBERTa classifiers, they find that bias mitigation does not consistently produce fairer models, and that fairness and explainability are orthogonal. Their explainability outcomes are rationale plausibility (human--model
alignment) and sufficiency (Area Over the Perturbation Curve, AOPC, where
higher is more faithful); across their fairness methods on ECtHR, AOPC
barely moves from the baseline ($77.4$, ranging $77.0$ to $78.8$).
This paper extends that line in three respects: it studies the \texttt{ecthr\_a} multi-label task rather than a single-label five-article subset; it operationalises fairness over applicant gender and ethnicity rather than defendant state; and it asks whether SCM-based debiasing changes the \emph{faithfulness} of post-hoc SHAP and Integrated Gradients~\citep{sundararajan2017axiomatic} explanations, which their design does not address.

Other work points the same way from different angles. \citet{yang2024fair} argue that, under moderate assumptions, controlling for fairness can influence feature attributions and vice versa, and find that optimising distributive fairness reduces explanation divergence across groups even without targeting it. \citet{wang2025bridging} find in hate-speech detection that input-based explanations can flag biased predictions but are unreliable for selecting fair models, reinforcing that explanations are an unreliable fairness-auditing tool. \citet{zhou2022explainability} argue explanations should be evaluated for whether they expose or conceal discriminatory patterns, not only for accuracy.

Taken together, these studies converge on a consistent picture: fairness and explainability do not automatically improve together, and their relationship is complex and context-dependent. No prior study has examined this relationship for SCM-based debiasing, nor measured the faithfulness, as opposed to the plausibility, of post-hoc explanations. This paper addresses that gap by jointly evaluating SCM-based fairness regularisation and SHAP explanation faithfulness on the ECtHR corpus.

\subsection{The Stereotype Content Model as a Basis for Debiasing}
\label{sec:rw-scm}

Most debiasing work in NLP targets specific demographic axes, typically gender or race, by contrasting word pairs such as ``man'' and ``woman'' \citep{bolukbasi2016debias}. This works for well-represented groups but breaks down for less-studied groups and for stereotyping that does not map onto explicit demographic markers, treating bias as a surface phenomenon when it is rooted in broader patterns of how social groups are perceived.

The Stereotype Content Model (SCM) offers a different starting point \citep{fiske2002model}: rather than cataloguing bias group by group, it holds that stereotypes are organised around two dimensions, warmth (perceived trustworthiness and intent) and competence (perceived capability), with marginalised groups positioned as either cold but competent or warm but incompetent across many social and cultural contexts \citep{fiske2018stereotype}. These dimensions are geometrically encoded in embedding spaces: \citet{fraser2021understanding} define warmth and competence axes in semantic embedding space and show that the resulting four quadrants recover documented group stereotypes, indicating the structure is a measurable property of learned representations. Building on this, \citet{ungless2022robust} and \citet{omrani2023scm} develop debiasing methods that operate directly on warmth--competence structure. More recent work extends SCM-based mitigation across representation and task levels. \citet{garbat2026warmth} operationalise warmth and competence as explicit directions in contextual BERT embeddings and apply projection-based mitigation, while \citet{zhu2025taskadaptive} use SCM scores to adapt adversarial debiasing strength at the instance level during sentiment classification. \citet{kocadag2026assessing} extend SCM mitigation to intersectional gender--race representations in static embeddings and evaluate how debiasing affects semantic coherence. Together, these studies show that SCM can support both representation-level and downstream-task interventions. The present work differs by applying an SCM-inspired contrastive margin regulariser during LegalBERT fine-tuning and by jointly evaluating downstream demographic fairness and post-hoc explanation faithfulness.

\subsection{Explainability and Faithfulness in NLP}
\label{sec:rw-faith}

When models are used in consequential settings such as law or healthcare, why a model made a prediction matters as much as whether it was correct, driving interest in post-hoc explanation methods such as SHAP \citep{lundberg2017unified} and LIME \citep{ribeiro2016lime}. A central problem is that a convincing explanation need not be an accurate one. \citet{jacovi2020towards} distinguish plausibility, appearing reasonable to a human reader, from faithfulness, reflecting the model's internal reasoning; the two do not always align, and \citet{atanasova2023faithfulnesstestsnaturallanguage} show empirically that coherent explanations can misrepresent the model's true decision process. In legal contexts this is especially concerning: an unfaithful explanation can project principled reasoning while concealing biased or arbitrary patterns.

A growing body of work therefore assesses the faithfulness of post-hoc explanations \citep{lyu2022generating}. Following the ERASER benchmark \citep{deyoung2020eraser}, faithfulness is decomposed into sufficiency, whether the highest-scored tokens reproduce the prediction, and comprehensiveness, whether all indicative tokens receive high scores; perturbation-based evaluation operationalises these by masking the features an explanation deems important and tracking the change in prediction \citep{deyoung2020eraser}. These are the metrics adopted here, and \citet{brandl2024interplay} use the same sufficiency criterion via AOPC, making the present results directly comparable in kind. \citet{manerba2022investigating} examine the effect of debiasing on classification performance and explanations, but do not investigate whether debiasing affects their \emph{faithfulness}, the specific question pursued here.

\subsection{Fairness Definitions and Metrics}
\label{sec:rw-fairness}

The normative basis for fairness in judicial prediction is equal treatment: under ECHR Article~14 \citep{chalkidis2022fairlex}, a model should not produce systematically different predictions for legally equivalent cases differing only in the parties' demographic characteristics. The mechanism is typically indirect; models learn semantic correlations between demographic language cues and outcomes from corpora encoding prior institutional disparities \citep{caliskan2017semantics, gallegos2024survey}.

Group fairness metrics capture systematic differences across socially defined groups and are appropriate here \citep{pessach2022fairness}. This paper reports three complementary metrics: Demographic Parity Difference (DPD), measuring unequal positive-prediction rates; Disparate Impact (DI), the ratio of positive-prediction rates under the four-fifths rule; and Equalized Odds Difference (EOD), which conditions on the ground-truth label to capture whether errors are distributed equitably. Reporting more than one is necessary because they can diverge: a model may satisfy parity while producing unequal error rates \citep{chouldechova2017fair}. Recent multi-label NLP evidence further shows that aggregate scores can conceal substantial label-level disparities, motivating class-conditional fairness reporting \citep{naranbat2025fairnessmetric}. Following \citet{cao2021}, fairness is evaluated with extrinsic task-level metrics rather than intrinsic ones, which frequently fail to predict downstream fairness. Because ECHR cases are organised by article, fairness is assessed conditionally on article type to detect localised disparities \citep{foulds2019intersectionaldefinitionfairness}, with a reduction of $\Delta\text{DPD} \geq 0.05$ pre-registered as the threshold for a practically meaningful improvement \citep{caton2024fairness}. For legal NLP specifically, \citet{chalkidis2022fairlex} introduce the FairLex benchmark for fairness in legal text processing, including ECtHR, situating the present evaluation within an established framework. 

\section{Methodology}
\label{sec:methodology}
 
\begin{figure*}[t]
  \centering
  \includegraphics[width=0.8\textwidth]{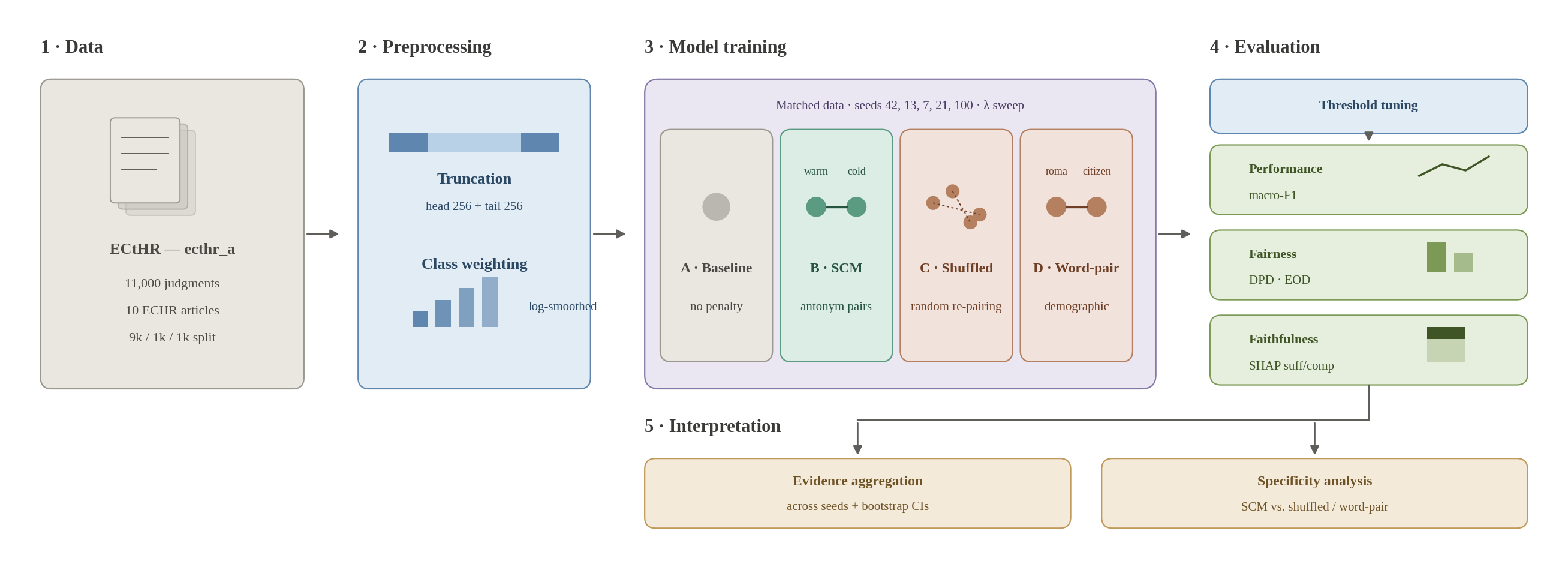}
  \caption{Experimental pipeline. The ECtHR \texttt{ecthr\_a} corpus is
  tokenised and class-weighted, then used to train four matched conditions
  under identical seeds (42, 13, 7, 21, 100) and a $\lambda$ sweep:
  (A) baseline, (B) SCM antonym pairs, (C) shuffled-pair control, and
  (D) demographic word-pair control. All are evaluated after threshold
  tuning along performance, fairness, and faithfulness, then compared across
  seeds and against the two controls.}
  \label{fig:pipeline}
\end{figure*}

This section describes the experimental design, dataset, models, and
evaluation framework used to test whether SCM-inspired contrastive
regularisation affects predictive performance, demographic fairness,
and explanation faithfulness in legal document classification. The
pipeline is shown in Figure~\ref{fig:pipeline}; the technical details
of the regularised models and controls are given in
Section~\ref{sec:scm-model} and Appendix~\ref{app:scm-pairs}.

\subsection{Research Design}

This research employs a controlled experimental design. The independent
variable is the regularisation applied during fine-tuning. The dependent variables are predictive performance (macro F1), fairness
(DPD and EOD, with DI diagnostic), and explanation faithfulness
(perturbation-based sufficiency and comprehensiveness).

The experiment compares four matched conditions: an unregularised
LegalBERT baseline, an SCM-inspired contrastive regularised model, a
shuffled-pair control, and a demographic word-pair debiasing control.
The two controls test, respectively, whether observed effects are
specific to the SCM warmth--competence structure or also arise under a
conventional protected-attribute word-pair formulation
(Section~\ref{sec:scm-model}). All four
conditions are trained and evaluated under identical data, tokeniser,
truncation, optimiser, and random seeds (42, 13, 7, 21, 100), so that
any observed difference is attributable to the regularisation term, and
all results are reported as mean and standard deviation across seeds.

The study is structured around three pre-registered hypotheses,
formulated prior to running any experiments on the basis of the
exploratory data analysis and prior literature. H1
(fairness--performance trade-off) holds that SCM-based regularisation
improves at least one fairness metric across the majority of reliable
articles, at the cost of a statistically significant but practically
small reduction in macro~F1 ($\Delta$F1 $< 0.03$). H2 (faithfulness
effect) holds that SCM regularisation changes the faithfulness of
post-hoc explanations relative to the baseline, with the direction of
that change an open empirical question. H3 ($\lambda$ trade-off) holds
that higher $\lambda$ produces a monotonic trade-off between fairness
and classification performance, while the relationship between
$\lambda$ and faithfulness is non-monotonic.

\subsection{Dataset}

The analysis uses the ECtHR alleged-violations corpus from the LexGLUE
benchmark (configuration \texttt{ecthr\_a})~\cite{chalkidis2022}, which
contains 11{,}000 European Court of Human Rights judgments. Each case
is a sequence of fact paragraphs with binary labels for ten ECHR
articles, making the task multi-label classification. The corpus suits
this study for three reasons: its fixed, publicly available
human-annotated labels support reproducible evaluation~\cite{chalkidis2022};
its multi-label structure with severe class imbalance (a 114.7$\times$
ratio between the most and least frequent labels) is a realistic legal
NLP setting; and its lack of explicit demographic labels makes it an apt
testbed for social-group-agnostic debiasing, which would otherwise
require external annotation. No missing values were found, and the
dataset is pinned to its download version. The data follow the official
LexGLUE split: training (9{,}000, 81.8\%), validation (1{,}000, 9.1\%),
and test (1{,}000, 9.1\%). Table~\ref{tab:label_freq} reports per-article
label frequencies.

\begin{table}[h] \caption{Label frequency per ECHR article across dataset splits.} \label{tab:label_freq} \begin{tabular}{lrrrr} \toprule \textbf{Article} & \textbf{Train} & \textbf{Train \%} & \textbf{Val} & \textbf{Test} \\ \midrule Art.6 & 4{,}704 & 52.3 & 300 & 299 \\ P1-1 & 1{,}421 & 15.8 & 139 & 122 \\ Art.5 & 1{,}368 & 15.2 & 187 & 166 \\ Art.3 & 1{,}349 & 15.0 & 193 & 189 \\ Art.8 & 710 & 7.9 & 87 & 123 \\ Art.2 & 505 & 5.6 & 57 & 56 \\ Art.10 & 291 & 3.2 & 42 & 77 \\ Art.14 & 141 & 1.6 & 18 & 16 \\ Art.11 & 110 & 1.2 & 33 & 37 \\ Art.9 & 41 & 0.5 & 4 & 5 \\ \bottomrule \end{tabular} \end{table}

This imbalance, together with 914 training cases (10.2\%) that carry no
violation label, motivates the class-weighted loss and requires fairness
evaluation to flag under-sampled articles as unreliable
(Section~\ref{sec:eval}). The documents are also long relative to the
input limit: the median training document is 984~words (mean~1{,}619;
max~35{,}416), and roughly 79.6\% exceed the 512-token limit. A
\textit{head\,+\,tail} truncation strategy retains the first 256 and last
256 subword tokens, preserving opening procedural context and closing
legal reasoning, which outperforms head-only truncation on long legal
documents~\cite{sun2019}.

\subsection{Baseline Model}

The primary baseline is \textsc{Legal-BERT-Base-Uncased}~\cite{chalkidis2020legalbert}
(\texttt{nlpaueb/\allowbreak legal-\allowbreak bert-\allowbreak base-\allowbreak uncased}),
a domain-adapted BERT variant pretrained on 12\,GB of English legal text.
A linear multi-label classification head is appended to the
\texttt{[CLS]} representation, followed by a sigmoid activation. The model
is fine-tuned with a Binary Cross-Entropy loss weighted by log-smoothed
inverse-frequency class weights:

\begin{equation}
  w_i = \frac{\log\!\left(1 + \frac{N^-_i}{N^+_i}\right)}
             {\operatorname{mean}_j\!\left[\log\!\left(1 +
             \frac{N^-_j}{N^+_j}\right)\right]}
  \label{eq:weights}
\end{equation}

\noindent
where $N^+_i$ and $N^-_i$ are the numbers of positive and negative
training examples for article~$i$, and log-smoothing prevents the extreme
114.7$\times$ imbalance ratio from destabilising training. Training
follows the hyperparameter recommendations of \citet{chalkidis2022}: a
learning rate of $2\times10^{-5}$, the AdamW optimiser with weight
decay~$0.01$, a batch size of~8, and early stopping on validation
macro~F1 with patience~3. After training, per-article thresholds $\tau \in [0.05, 0.95]$ are tuned on
the validation set to maximise F1. This matters most for minority articles,
where a fixed 0.5 threshold yields near-zero recall, and ensures the baseline
is not handicapped by poor thresholding.

\subsection{Fairness-Regularised Model}
\label{sec:scm-model}

The experimental condition augments fine-tuning with SCM-based
regularisation inspired by \citet{omrani2023scm}. Rather than specifying
demographic word pairs, it targets stereotype-relevant structure along
the SCM dimensions of \textit{warmth} and \textit{competence} using
antonym pairs: 19 per dimension (14 from \citet{omrani2023scm} and 5 from
the warmth--competence lexicon of \citet{Ayesh2026AnnotatingDO} on each
axis), 38 in total (Appendix~\ref{app:scm-pairs}). This suits legal NLP because
ECtHR documents carry no explicit demographic labels, and because it
targets the implicit associations through which bias manifests in legal
language rather than requiring the model to detect demographic markers.

During fine-tuning, an SCM term is added to the classification loss:

\begin{equation}
  \mathcal{L}_{\text{total}} =
      \mathcal{L}_{\text{CE}} + \lambda \cdot \mathcal{L}_{\text{SCM}}
  \label{eq:loss}
\end{equation}

\noindent
where $\mathcal{L}_{\text{CE}}$ is the weighted loss of
Equation~\ref{eq:weights}, $\lambda$ controls regularisation strength,
and $\mathcal{L}_{\text{SCM}}$ is a contrastive margin loss over the
antonym pairs present in each batch:

\begin{equation}
  \mathcal{L}_{\text{SCM}} =
      \frac{1}{|\mathcal{P}|}
      \sum_{(w^+,\, w^-)\,\in\,\mathcal{P}}
      \max\!\Bigl(0,\;
        \cos\!\bigl(\mathbf{h}(w^+),\, \mathbf{h}(w^-)\bigr)
        + m
      \Bigr)
  \label{eq:scm}
\end{equation}

\noindent
where $\mathcal{P}$ is the set of pairs in the batch,
$\mathbf{h}(\cdot)$ is the final-layer contextual embedding,
$\cos(\cdot,\cdot)$ is cosine similarity, and $m = 0.5$ is the margin.
Every pair contributes equally, and the loss is zero only when a pair
reaches $\cos \leq -m$ (strong opposition); otherwise it is positive and
the gradient pushes the pair further apart.

This should be read as an SCM-inspired representational constraint, not a
direct implementation of projection-based debiasing: it constrains the
geometry of warmth--competence pairs but does not remove associations
between demographic terms and SCM traits, so the intervention may alter
representational structure without reducing downstream prediction
disparity. The margin formulation was adopted because a plain cosine loss saturated after
the first epoch; the margin clamps the loss at zero, keeping the signal on
pairs that still need separation. The strength
$\lambda$ is tuned on validation macro~F1 over
$\{0.01,\, 0.05,\, 0.1,\, 0.5\}$. This uniform formulation is the main
intervention; a variant weighting each document by its stereotypical
content is examined in Appendix~\ref{app:iw}.

\subsection{Control Conditions}
\label{sec:controls}

Two controls separate the SCM warmth--competence structure from contrastive
regularisation in general. The \emph{shuffled-pair control} keeps the SCM
term fixed except for the structure it encodes: the same antonym vocabulary
is used, but each word is paired with a random partner, so the pairs no
longer express the warmth--competence opposition. Pair count, contrastive
loss (Equation~\ref{eq:scm}), margin ($m = 0.5$), $\lambda$ grid, seeds, and the separately trained baseline are identical in construction to the SCM condition. If the SCM effects depend on the antonym
structure they should disappear here; if they follow from contrastive
regularisation of any kind they should persist.

The \emph{demographic word-pair control} instead tests whether the
conventional paradigm that contrasts explicit protected-attribute
terms~\cite{bolukbasi2016debias} would move disparity where the SCM term did
not. It reuses the identical loss, margin, and $\lambda$ grid
(Equations~\ref{eq:loss}--\ref{eq:scm}), substituting demographic pairs
$\mathcal{P}$ for the antonyms, with each axis debiased separately against
its own freshly trained baseline under the same five seeds. The gender axis
uses nineteen genuine binary opposites (\emph{man}--\emph{woman},
\emph{he}--\emph{she}, and so on), the canonical target of word-pair
debiasing. The ethnicity axis has no natural antonyms, so each of sixteen
minority terms is contrasted with an unmarked majority-referent term
(e.g.\ citizen, national, resident); this minority--majority
construction is deliberately artificial, included only to test whether the
approach applies to that axis at all. Results are in
Appendix~\ref{sec:wordpair-results}.

\subsection{Evaluation Framework}
\label{sec:eval}

Predictive performance is assessed using macro-averaged F1 on the
held-out test set; macro~F1 is preferred over micro~F1 because it weights
all articles equally and is therefore sensitive to minority classes.
Throughout, we avoid over-reliance on binary significance decisions,
reporting per-seed differences, their direction and bootstrap confidence
intervals rather than $p$-values alone, since a signed-rank
test cannot reach significance at five seeds.

Fairness is evaluated using extrinsic task-level metrics following
\citet{cao2021}, who show that intrinsic representation-level metrics
frequently fail to predict downstream task-level fairness. Protected
groups are operationalised through the presence of demographic keywords in
the document text, matched using whole-word regular expressions, and
defined separately for gender (40 keywords) and ethnicity (23 keywords).
The ethnicity term list is informed by the demographic lexicon of
\citet{choenni2021}, as adapted in
\citet{kocadag2026assessing}, but is restricted to ethnicity and
minority-status terms and supplemented with groups salient in ECtHR
proceedings such as Roma, Kurdish, Chechen, asylum-seeker, and refugee.
Nationality adjectives (e.g.\ European, American, British) were excluded:
respondent states appear in nearly every judgment, so including them would
assign half the corpus to the protected group and dissolve the comparison. The full keyword lists are provided in
Appendix~\ref{app:keywords}. Under this operationalisation the test set
splits into 931 protected and 69 unprotected documents for gender, and
136 protected and 864 unprotected documents for ethnicity; the severe
imbalance of the gender split is a consequence of the keyword-based
operationalisation rather than of the dataset itself, and its
implications are discussed in Section~\ref{sec:discussion}. To test
whether the fairness conclusions depend on this keyword operationalisation,
the ethnicity grouping is additionally computed with a named-entity
recogniser (spaCy \texttt{en\_core\_web\_sm}), which assigns a document to
the protected group when it contains a recognised entity span matching a
minority or ethnic group rather than a bare string match; this yields a
smaller, more conservative protected group (59 versus 136 documents) and
serves as a robustness check, reported in
Section~\ref{sec:results} and Appendix~\ref{app:fairness-ci}.

The primary fairness metric is the Demographic Parity Difference (DPD),
the absolute difference in positive prediction rates across groups, where
0.0 indicates perfect parity. Equalized Odds Difference (EOD) conditions
on the ground-truth label and is reported as the larger of the per-group
true-positive-rate gap and false-positive-rate gap, so a higher value
indicates a less equitable distribution of errors. Disparate Impact (DI),
the ratio of positive prediction rates between groups, is reported
diagnostically in Appendix~\ref{app:fairness-ci} (Table~\ref{tab:di});
DI is unstable when a group has few positive predictions, so DPD carries the primary conclusions. A practically meaningful improvement is pre-registered
as a reduction of $\Delta\text{DPD} \geq 0.05$. Articles with more than 37
ground-truth positive labels in the test set are treated as reliable,
yielding seven reliable articles (Art.2, 3, 5, 6, 8, 10, P1-1); Art.11
(exactly 37), Art.14 (16), and Art.9 (5) fall below this and are excluded
from the primary conclusions. This threshold was set by bootstrap variance
analysis of the EDA results, and bootstrap confidence intervals (95\%,
$n = 10{,}000$) for the flagged articles are reported in
Appendix~\ref{app:fairness-ci}. The same seven articles are used for the
faithfulness macro-average, so the two analyses are reported over a
consistent article set.

Explanation faithfulness is assessed using perturbation-based evaluation, following the sufficiency and comprehensiveness definitions of the ERASER benchmark~\cite{deyoung2020eraser}.
SHAP values~\cite{lundberg2017unified} are computed for both models using
KernelSHAP with 10 background samples and 128 perturbation samples per
instance, on a random sample of 200 test instances. The small background set trades off the high cost of KernelSHAP on 512-token
inputs; to check the sufficiency results are not an artefact of this choice,
the analysis is repeated with Integrated Gradients~\citep{sundararajan2017axiomatic}, which needs no background
sampling, with cross-method agreement across all five seeds as the criterion
for a consistent effect. Two
metrics are computed for $k \in \{1\%,\, 5\%,\, 10\%\}$ of input tokens.
\emph{Sufficiency} is the magnitude of the change in prediction
probability when only the top-$k$ attributed tokens are retained and all
others are replaced with \texttt{[MASK]}; a \emph{lower} value indicates a
more faithful explanation, since the retained tokens alone nearly
reproduce the original prediction. \emph{Comprehensiveness} is the change
in prediction probability when the top-$k$ tokens are removed; a
\emph{higher} value indicates a more faithful explanation, since the
removed tokens were genuinely important. The two metrics therefore have
opposite desirable directions, and the change is measured as an absolute
difference, capturing the magnitude of the probability shift rather than
its sign.

\subsection{Experimental Procedure and Pre-specified Decisions}
\label{sec:procedure}

The experimental procedure comprises four sequential stages. First, the
unregularised LegalBERT baseline is trained and evaluated to establish the
reference performance, fairness, and explanation-faithfulness levels.
Second, the regularisation strength $\lambda$ is selected using validation
macro~F1. Third, the selected regularised model is compared with the
baseline across predictive performance, demographic fairness, and
explanation faithfulness, with uncertainty characterised through bootstrap
confidence intervals and consistency assessed across random seeds. Finally,
a sensitivity analysis examines how these outcomes vary across the tested
values of $\lambda$. Because full attribution-based faithfulness evaluation
is computationally expensive, the $\lambda$ sweep for faithfulness is
performed for a single seed, whereas the corresponding fairness analysis is
aggregated across all five seeds.
 
 

\section{Results}
\label{sec:results}

This section reports the empirical findings of the controlled experiment. All
results are reported on the held-out test set ($n=1{,}000$) and aggregated over
five random seeds (42, 13, 7, 21, 100) unless stated otherwise. Reporting over
multiple seeds captures variation from training randomness, not only test-set
sampling noise; single-seed estimates are used only where explicitly labelled
as illustrative. Values are reported as mean $\pm$ standard deviation across
seeds. Fairness and faithfulness are analysed over the seven reliable articles
defined in Section~\ref{sec:eval}.

Table~\ref{tab:performance} reports macro-averaged F1 for the baseline and the
SCM model, where each seed's SCM result uses the regularisation strength
$\lambda$ selected on that seed's validation set (per-seed
$\lambda \in \{0.1,0.05,0.01,0.5,0.5\}$). Averaged over seeds, the baseline
reaches $0.670 \pm 0.013$ and the SCM model $0.656 \pm 0.012$, a mean
difference of $-0.014$. The per-seed differences are $+0.014$, $-0.026$,
$-0.028$, $-0.008$, and $-0.024$: four of the five seeds favour the baseline,
and only seed 42 favours SCM. The two distributions overlap substantially
across seeds.

\begin{table}[H]
\centering
\caption{Test macro F1 over five seeds (mean $\pm$ std). SCM uses each seed's
validation-selected $\lambda$.}
\label{tab:performance}
\small
\begin{tabular}{lc}
\toprule
Model & Macro F1 \\
\midrule
Baseline (LegalBERT) & $0.670 \pm 0.013$ \\
SCM (per-seed $\lambda$) & $0.656 \pm 0.012$ \\
$\Delta$ & $-0.014$ \\
\bottomrule
\end{tabular}
\end{table}

This revises the single-seed picture. On seed 42 alone, SCM appeared to improve
test F1 by roughly $+0.013$; averaged over five seeds this advantage disappears
and reverses to a small negative point estimate. The single-seed gain was within
the range of seed-to-seed variation and did not reflect a systematic effect.
SCM regularisation therefore does not improve classification performance, and
any effect on F1 is small and within training noise. H1's expected performance
cost is likewise not observed: SCM neither reliably helps nor reliably harms
macro F1. The seed-42 validation grid illustrates the non-monotonic dependence
on $\lambda$ (validation F1 peaks at $\lambda=0.1$ with 0.683, falling to
0.665, 0.669, and 0.653 at the other values); this motivates per-seed
$\lambda$ selection rather than a performance claim.

To complete the characterisation of the baseline, its explanation faithfulness
is reported alongside the SCM model in Table~\ref{tab:faithfulness}. Averaged
over five seeds, the baseline reaches SHAP sufficiency of
$0.1289 \pm 0.0108$ at $k=1\%$, providing the reference point against which the
regularised condition is compared.

Turning to fairness, with protected groups defined as in Section~\ref{sec:eval},
both axes are imbalanced in opposite directions. The gender split has a small
unprotected group, while the ethnicity split has a small protected group. The
fairness evaluation therefore rests on per-article reliability filtering and
bootstrap confidence intervals rather than on group balance.
Table~\ref{tab:fairness_multiseed} reports mean absolute DPD and mean EOD over
the reliable articles, averaged across seeds. On both axes the SCM model remains
within seed-to-seed variation of the baseline. Mean $|\mathrm{DPD}|$ is
essentially unchanged on ethnicity ($0.1188 \pm 0.0105 \rightarrow
0.1166 \pm 0.0026$) and slightly higher on gender ($0.1490 \pm 0.0102
\rightarrow 0.1555 \pm 0.0056$); in neither case does the change indicate a
consistent reduction in disparity. Mean EOD moves in opposite directions on the
two axes and likewise shows no consistent improvement.

\begin{table}[H]
\centering
\caption{Fairness over the seven reliable articles, mean $\pm$ std across five
seeds. Lower $|\mathrm{DPD}|$ and lower EOD indicate less disparity. No metric
shows a consistent improvement under SCM.}
\label{tab:fairness_multiseed}
\small
\begin{tabular}{llcc}
\toprule
Axis & Metric & Baseline & SCM \\
\midrule
Gender & mean$|\mathrm{DPD}|$ & $0.1490 \pm 0.0102$ & $0.1555 \pm 0.0056$ \\
       & mean EOD & $0.2162 \pm 0.0388$ & $0.1934 \pm 0.0194$ \\
Ethnicity & mean$|\mathrm{DPD}|$ & $0.1188 \pm 0.0105$ & $0.1166 \pm 0.0026$ \\
          & mean EOD & $0.1851 \pm 0.0128$ & $0.2025 \pm 0.0187$ \\
\bottomrule
\end{tabular}
\end{table}

The per-article evidence reinforces this null. For each seed, the analysis counts
how many of the seven reliable articles have a paired-bootstrap $\Delta$DPD
confidence interval that crosses zero. On gender this count is 6, 6, 6, 5, and 5
across the five seeds (mean 5.6 of 7); on ethnicity it is 6, 7, 5, 5, and 6
(mean 5.8 of 7). In the large majority of article--seed combinations, the
change in disparity is indistinguishable from zero. The few articles whose
intervals exclude zero do not share a consistent sign or location across seeds.
An independent named-entity-based ethnicity grouping reproduces the pattern: a
mean of 5.0 of 7 reliable articles cross zero across seeds, compared with 5.8
under keyword matching. Full per-article intervals for both operationalisations
are reported in Appendix~\ref{app:fairness-ci}.

Disparate Impact is reported in the appendix as a supporting metric only. It is
unstable under the group imbalance on both axes, and its across-seed standard
deviation is large relative to the effect, so the primary conclusions rely on
DPD and EOD. H1's fairness component is therefore not confirmed: SCM
regularisation does not produce a consistent or practically meaningful reduction
in demographic disparity on either axis under this operationalisation. The
demographic word-pair control reaches the same null on the gender axis
(Appendix~\ref{app:wordpair}).

Table~\ref{tab:faithfulness} reports SHAP and Integrated-Gradients sufficiency
for the baseline and SCM models at $k \in \{1\%,5\%,10\%\}$, averaged over five
seeds. Lower sufficiency indicates greater faithfulness. SCM regularisation
increases SHAP sufficiency at every value of $k$, and the increase grows with
$k$ (from approximately $+0.031$ at $k=1\%$ to $+0.053$ at $k=10\%$). The
standard-deviation bands for baseline and SCM SHAP sufficiency do not overlap at
any $k$. Integrated Gradients agrees in direction at all three thresholds.
SHAP comprehensiveness rises only slightly under SCM, with overlapping bands,
and is therefore not interpreted as a meaningful change.

\begin{table}[H]
\centering
\caption{Faithfulness over five seeds (mean $\pm$ std),
$k \in \{1\%,5\%,10\%\}$. Lower sufficiency is more faithful. SCM raises SHAP
sufficiency at every $k$ with non-overlapping bands; Integrated Gradients agrees
in direction.}
\label{tab:faithfulness}
\scriptsize
\setlength{\tabcolsep}{2.8pt}
\begin{tabular}{c@{\hspace{4pt}}cc@{\hspace{6pt}}cc}
\toprule
& \multicolumn{2}{c}{SHAP Suff.} & \multicolumn{2}{c}{IG Suff.} \\
\cmidrule(lr){2-3}\cmidrule(lr){4-5}
$k$ & Base & SCM & Base & SCM \\
\midrule
$1\%$  & $0.1289 \pm 0.0108$ & $0.1601 \pm 0.0166$ & $0.1284 \pm 0.0103$ & $0.1677 \pm 0.0222$ \\
$5\%$  & $0.1282 \pm 0.0108$ & $0.1697 \pm 0.0159$ & $0.1264 \pm 0.0101$ & $0.1952 \pm 0.0375$ \\
$10\%$ & $0.1277 \pm 0.0110$ & $0.1803 \pm 0.0184$ & $0.1237 \pm 0.0101$ & $0.2572 \pm 0.0808$ \\
\bottomrule
\end{tabular}
\end{table}

The direction of the sufficiency effect holds in all five seeds at all three
values of $k$. With only five seeds, a signed-rank test cannot return a
$p$-value below 0.0625, so the evidence is reported as cross-seed directional
consistency rather than as a single significance test. H2 is supported in a
specific form: the change is a degradation concentrated in sufficiency. The
cross-encoder replication on general-domain BERT and RoBERTa provides supporting
evidence that the effect is not unique to LegalBERT; full results are reported
in Appendix~\ref{app:cross-encoder}.

To test whether these effects are specific to the SCM contrastive structure or
follow from contrastive regularisation more generally, the full pipeline was
repeated with shuffled pairs. Under shuffled-pair regularisation, macro F1 is
$0.661 \pm 0.012$ for the baseline and $0.657 \pm 0.014$ for the regularised
model (mean difference $-0.004$). Mean $|\mathrm{DPD}|$ is also flat: gender
$0.1573 \pm 0.0054 \rightarrow 0.1508 \pm 0.0127$ and ethnicity
$0.1205 \pm 0.0056 \rightarrow 0.1208 \pm 0.0071$. SHAP sufficiency at
$k=5\%$ rises from $0.1228 \pm 0.0064$ to $0.1618 \pm 0.0070$ (mean increase
$+0.039$, present in all five seeds). Thus the control reproduces the three main
patterns: flat performance, flat fairness, and degraded sufficiency. The effect
is therefore attributable to contrastive representational regularisation of
this form rather than specifically to the warmth--competence pairing.

Finally, across the five-seed $\lambda$ sweep, mean $|\mathrm{DPD}|$ over the
seven reliable articles remains flat and never improves on the baseline. For
gender, the baseline is 0.1490 compared with 0.1575, 0.1592, 0.1551, and 0.1518;
for ethnicity, the baseline is 0.1188 compared with 0.1206, 0.1199, 0.1196, and
0.1209. No $\lambda$ approaches the pre-registered
$\Delta\mathrm{DPD} \geq 0.05$ threshold. The full fairness sweep is reported
in Appendix~\ref{app:lambda}; the single-seed faithfulness sweep there is
non-monotonic. H3 is therefore only partially supported: the expected
fairness--performance trade-off is absent, while the illustrative faithfulness
curve is non-monotonic.

\FloatBarrier

\section{Discussion}
\label{sec:discussion}

This section interprets the findings against prior work, considers alternative explanations, and reflects on limitations. The central result is a five-seed dissociation: SCM regularisation leaves demographic disparity and classification performance largely unchanged, while consistently degrading explanation faithfulness.

H1 anticipated a small performance cost alongside a fairness gain; it is rejected on both components, as the paragraphs below show (Table~\ref{tab:performance}). The earlier single-seed result, which
suggested a small SCM advantage, was an artefact of seed selection and reverses
under multi-seed evaluation. The most defensible reading is that SCM
regularisation leaves classification performance essentially intact, consistent
with Brandl et al.~\cite{brandl2024interplay}, who find that fairness
interventions on ECtHR do not consistently degrade performance. The single-seed
$\lambda$ grid further shows that only the strongest regularisation
($\lambda = 0.5$) lowers validation F1, so the absence of a performance effect
is specific to moderate regularisation rather than a general property of the
loss. The practical implication is that the SCM term is performance-neutral,
which is what makes the faithfulness degradation interpretable: the explanation
effects cannot be attributed to a model that has simply become weaker at the
task.

The central finding on H1's fairness component is a null: SCM regularisation
does not produce a consistent or practically meaningful reduction in demographic
disparity. This null holds across two distinct fairness definitions,
parity-based and label-conditioned, rather than depending on a single metric.
It is established under a keyword-based operationalisation of the protected
groups, and should be read with that proxy in mind. The result is consistent
with a broad strand of the debiasing literature: Gonen and
Goldberg~\cite{gonen2019lipstick} show that many methods suppress surface
indicators of bias without removing its underlying geometric structure, and
Meade et al.~\cite{meade2022debiasingsurvey} find that no single method works
reliably across tasks and configurations. One likely reason is that the
contrastive loss operates on antonym-pair representations at the token level;
as ~\citet{kaneko2020notenough} observe,
representation-level debiasing does not consistently transfer to downstream
behaviour. The intervention may therefore reshape embedding geometry while
leaving prediction-level disparity intact. The better-balanced ethnicity axis (136 versus 864 documents) likewise shows no improvement: no $\lambda$ moves disparity toward the threshold, and a word-pair control is null even on gender; the faithfulness effect, by contrast, is detectable under the same seeds with non-overlapping bands, so the asymmetry is not an artefact of unequal power between the two measurements.

The comparison with Brandl et al.~\cite{brandl2024interplay} is instructive
because the two studies operationalise the protected group differently: they
binarise the defendant state, whereas this paper uses applicant gender and
ethnicity. Despite that difference, both reach a null on the same corpus. This
strengthens the suggestion that disparity in ECtHR predictions is driven by
factors that token-level interventions do not reach. The instance-weighted
variant reinforces this: concentrating regularisation on the most
stereotype-laden documents does not recover a fairness effect either. A
conventional demographic word-pair control reaches the same null even on the
gender axis where that paradigm is designed to apply, extending the null beyond
SCM regularisation to the dominant token-level debiasing approach.

The clearest finding concerns H2: SCM regularisation increases SHAP sufficiency
at every $k$, and because lower sufficiency is more faithful, this is a
degradation. The direction holds in all five seeds at all three $k$ without
exception, with non-overlapping bands. The effect is
also consistent under Integrated Gradients (Table~\ref{tab:faithfulness}),
which does not rely on background sampling; the agreement is clearest at the
smaller $k$ values, while at $k = 10\%$ the IG gap falls within the method's
noise band.

This contrasts with Brandl et al.~\cite{brandl2024interplay}, whose fairness
methods left AOPC sufficiency on ECtHR essentially unchanged. Their AOPC and
the ERASER sufficiency used here run in opposite directions, so the comparison
is one of movement, not level: their sufficiency did not move, whereas the
SCM intervention here does, towards less faithful explanations.
Comprehensiveness remains effectively flat with overlapping bands and is not
interpreted as a meaningful change. The asymmetry is informative: sufficiency degrades while comprehensiveness remains unchanged, so the intervention disrupts the top-token summary without changing the full attribution set’s contribution.  Whether this
reflects a substantive change in reasoning or a redistribution of attribution mass is left for future work with non-masking metrics.
The shuffled-pair control reframes the mechanism: because an arbitrary contrastive penalty reproduces the same effects, the contribution is better stated at the level of contrastive representational regularisation than of the SCM design alone. This does not weaken the dissociation, which concerns the fairness--faithfulness relationship rather than the SCM mechanism; if anything it broadens it, since the dissociation holds under shuffled-pair regularisation too.

The same intervention that fails to shift demographic disparity nonetheless
shifts explanation behaviour in a consistent, measurable way. The two outcomes
do not move together, which is the central finding. Its consequence for
legal-NLP auditing is that, in this setting, explanation faithfulness could not
have served as a proxy for fairness: a change in explanation behaviour carried
no information about whether the model had become fairer, in either direction.
An auditor watching only the explanation metrics would have inferred a
meaningful change while the fairness metrics did not move; fairness must
therefore be measured directly. This carries a direct ethical implication for
deployment: in high-stakes legal settings, a more coherent-looking explanation
could be mistaken for evidence of a fairer model. The null fairness result should likewise not be read as evidence
that the model is unbiased; it shows only that the token-level associations
targeted here are not the measurable source of disparity, while disparity
arising from how applicant groups are described across decades of court
documents may persist undetected. The result extends Brandl et
al.~\cite{brandl2024interplay}, who established orthogonality at the level of
rationale plausibility: moving to the stricter criterion of faithfulness shows
the orthogonality is asymmetric, one property unchanged while the other moves
consistently, which tightens the case that explanation quality and fairness are
independent outcomes in this setting.

H3 is partially confirmed. Within the single illustrative seed for which the
full grid was computed, sufficiency is non-monotonic in $\lambda$: it decreases
slightly at small $\lambda$, then rises above baseline at $\lambda = 0.1$ and is
worst at $\lambda = 0.5$. The monotonic fairness--performance trade-off H3 anticipated is not observed, because no tested value of $\lambda$ improved fairness in the five-seed $\lambda$ sweep
(Table~\ref{tab:apx:fairness-lambda}). Instead, the trade-off that appears is
between performance and local sufficiency. Because the selected $\lambda$ and
the curve differ across seeds, this single-seed grid illustrates only the shape
and is not used for an aggregate claim; the aggregate degradation already
reflects each seed's selected $\lambda$.

Several limitations qualify these conclusions. First, a construct-validity threat affects the fairness operationalisation. Protected groups are inferred from demographic
keywords, but ECtHR demographic terminology often functions as a legal concept
rather than a demographic identifier, so cases involving a group may not contain
the keywords and vice versa. This is most acute for gender, where the split is
severely imbalanced (931 versus 69). The fairness conclusions therefore rest
most strongly on the better-balanced ethnicity axis, the per-article reliability
filtering, and the named-entity robustness check. Because group membership is a
text-based proxy, the fairness null should be read as evidence about this
operationalisation, not as a claim that the model is unbiased with respect to
the underlying demographic groups.

Second, the faithfulness metrics carry a construct-validity limitation. Token-masking
sufficiency and comprehensiveness assume that model behaviour under masking
generalises to natural input variation. For a contextual model, replacing tokens
with \texttt{[MASK]} creates out-of-distribution inputs whose probability
estimates may be unreliable, so the sufficiency--comprehensiveness divergence
could partly reflect this artefact \cite{deyoung2020eraser}. The SHAP--IG agreement mitigates but does
not eliminate this concern, since both rely on perturbation-based reasoning.

Third, external validity is limited. All experiments use a
single corpus, language, and a small set of encoders. The cross-encoder runs
extend the faithfulness finding within the transformer family, but
generalisation to other jurisdictions, languages, or model families cannot be
assumed. The margin-based loss is also an SCM-inspired representational
constraint rather than a full projection-based debiasing method, and the
shuffled-pair control shows that its effects are not specific to the SCM
structure. Conclusions about ``SCM debiasing'' should therefore be read as
conclusions about contrastive representational regularisation of this form.

Finally, the regularisation strength $\lambda$ was selected on validation
macro~F1 rather than on a fairness criterion. A fairness-optimal or joint
selection might surface an effect that the performance-optimal $\lambda$ misses.
The per-$\lambda$ analysis argues against this, since no $\lambda$ improved
fairness, but it rests on a single seed. This is the most direct threat to the
fairness null and the clearest target for follow-up work.

The limitations translate into three directions for future work. First, the
fairness operationalisation should be strengthened. The named-entity-based
ethnicity grouping used here addresses operationalisation choice, but not the
deeper proxy problem, since both groupings still infer membership from text.
Replacing keyword and entity matching with structured case metadata would yield
better-balanced groups, particularly for gender, and is the single most
important open question left by this study. Second, the faithfulness analysis
should move beyond token masking to non-masking measures that do not create
out-of-distribution inputs. Third, the dissociation should be tested in a
setting where a fairness intervention actually changes fairness. The
demographic word-pair control rules out the obvious candidate method in this
setting, but pairing a method known to shift fairness with the faithfulness
evaluation used here remains necessary to establish whether the dissociation is
general or an artefact of an intervention that left fairness unchanged.

\section{Conclusion}
\label{sec:conclusion}
This paper asked to what extent SCM-based fairness regularisation influences
fairness and explanation faithfulness in legal document classification while
maintaining predictive performance. The answer is a dissociation: the same
intervention that fails to reduce demographic disparity consistently degrades
the faithfulness of post-hoc explanations, at no reliable cost to predictive
performance.

SCM regularisation does not produce a meaningful reduction in demographic
disparity. This null result is robust across two fairness definitions, the
more reliable ethnicity axis, an NER-based regrouping, an instance-weighted
variant, and a conventional demographic word-pair control on the gender axis
for which that approach is specifically designed. In contrast, SCM
regularisation consistently increases SHAP sufficiency at every threshold
$k$ across all five seeds, with non-overlapping standard-deviation bands and
the same directional effect observed with Integrated Gradients. At the same
time, comprehensiveness and classification performance remain essentially
unchanged.

Across the tested regularisation strengths, no value of $\lambda$ yields a
fairness improvement. The observed trade-off therefore concerns predictive
performance and local explanation sufficiency rather than fairness. Moreover,
the shuffled-pair control reproduces the same overall pattern, indicating that
neither the fairness null nor the degradation in explanation faithfulness is
specific to the SCM warmth--competence structure.

The contribution is to show that fairness and faithfulness are not coupled
outcomes of a debiasing intervention. Prior work established that fairness and
explainability are largely orthogonal at the level of rationale plausibility on
this corpus; this paper extends that to the stricter criterion of faithfulness
and shows the orthogonality is asymmetric, with one property unchanged while the
other moves consistently. The practical consequence is direct: in this setting,
a more coherent-looking explanation was not evidence of a fairer model, which
matters for how AI systems in high-stakes legal settings are audited.

These conclusions are qualified by the study's scope: one corpus, one language,
and a small set of transformer encoders; a keyword-based group operationalisation
that is an imperfect proxy, most acutely for the severely imbalanced gender axis;
and faithfulness metrics that rely on token masking, only partly mitigated by the
agreement between two attribution methods. The regularisation strength was also
selected on validation macro~F1 rather than on a fairness criterion, and the
per-$\lambda$ argument that no $\lambda$ would have helped rests on a single seed.
The fairness null should therefore be read as evidence that the token-level
associations SCM targets are not the primary source of measurable disparity under
this operationalisation, not as evidence that the model is unbiased. The most promising way forward is to retest the dissociation under a more reliable group definition and in a setting where a fairness intervention does move fairness. This would establish whether the dissociation is general or specific to an intervention that left fairness unchanged.

\bibliographystyle{ACM-Reference-Format}
\bibliography{bibtex_acm_updated}

\appendix

\section{Faithfulness and fairness across regularisation strength}
\label{app:lambda}
\begin{figure}[h]
  \centering
  \includegraphics[width=\columnwidth]{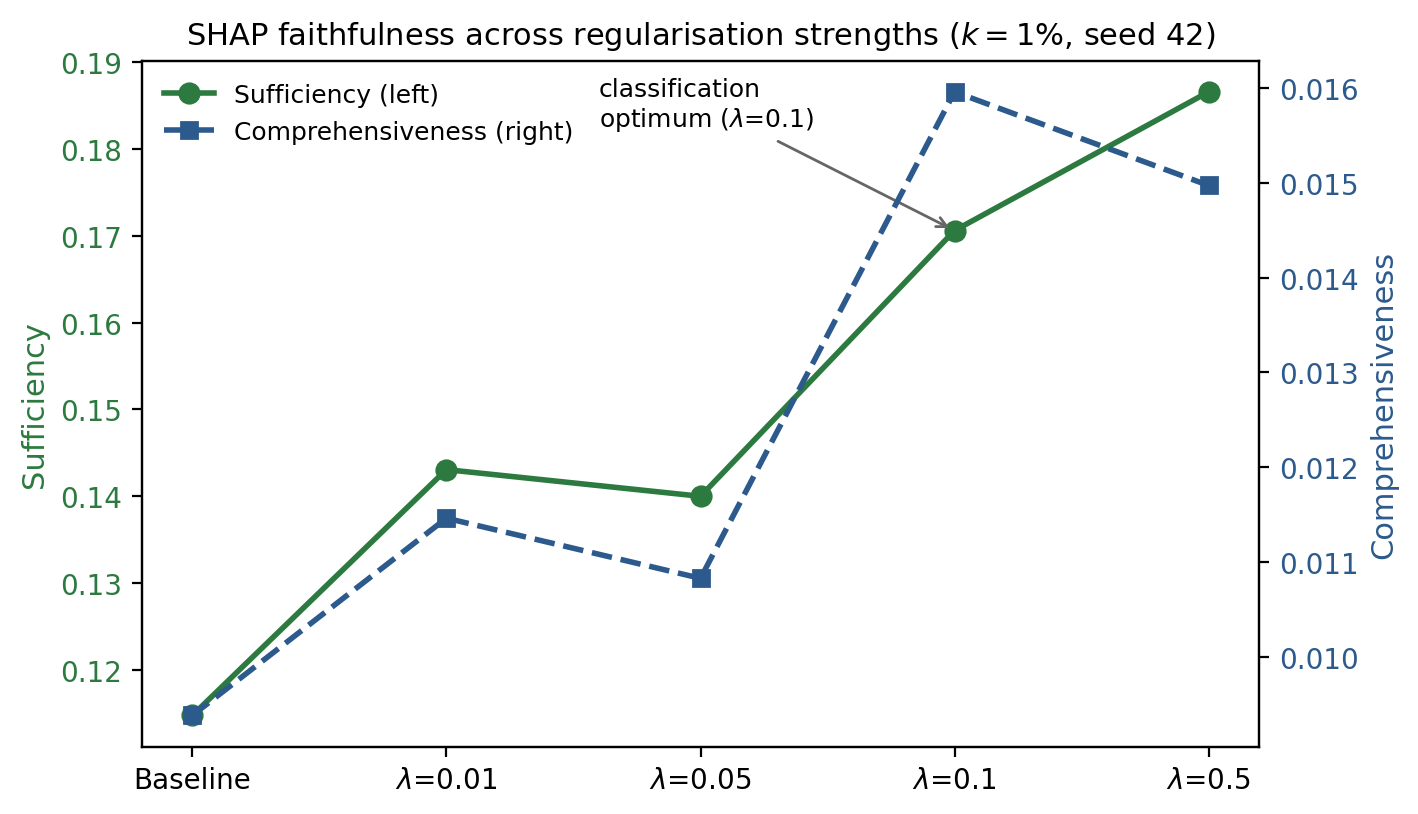}
  \caption{SHAP sufficiency (left axis) and comprehensiveness (right axis)
  across regularisation strengths at $k = 1\%$, single seed (seed 42). Lower
  sufficiency indicates a more faithful explanation. Sufficiency is lowest at
  the baseline, dips slightly at small $\lambda$, then rises monotonically to
  its maximum at $\lambda = 0.5$; $\lambda = 0.1$ is the optimum for
  classification performance, not for faithfulness. The curve illustrates the
  shape within one seed rather than an across-seed aggregate.}
  \label{fig:apx:faithfulness}
\end{figure}

This appendix presents the sensitivity analysis across the tested
regularisation strengths $\lambda$. The fairness sweep
(Table~\ref{tab:apx:fairness-lambda}) aggregates results over all five random
seeds, whereas the faithfulness sweep
(Figure~\ref{fig:apx:faithfulness}) is reported for seed~42 only. This
single-seed analysis is used to characterise the shape of the relationship
between regularisation strength and explanation faithfulness, since each
$\lambda$ value requires a separate full SHAP attribution run. The main-text
faithfulness results in Table~\ref{tab:faithfulness}, by contrast, are
aggregated across all five seeds and use the validation-selected $\lambda$
for each seed.

\begin{table}[h]
  \caption{Fairness across regularisation strengths $\lambda$, mean $\pm$ std
  over five seeds on the seven reliable articles. The baseline row is the
  unregularised model. Lower $|$DPD$|$ and lower EOD indicate less disparity.
  No value of $\lambda$ reduces disparity below the baseline on either axis,
  and none approaches the pre-registered $\Delta\text{DPD} \ge 0.05$
  improvement threshold.}
  \label{tab:apx:fairness-lambda}
  \begin{tabular}{llcc}
    \toprule
    Axis & $\lambda$ & mean$|$DPD$|$ & mean EOD \\
    \midrule
    Gender    & baseline & $0.1490 \pm 0.0102$ & $0.2162 \pm 0.0388$ \\
              & 0.01     & $0.1575 \pm 0.0097$ & $0.2148 \pm 0.0211$ \\
              & 0.05     & $0.1592 \pm 0.0085$ & $0.1926 \pm 0.0335$ \\
              & 0.1      & $0.1551 \pm 0.0097$ & $0.1929 \pm 0.0206$ \\
              & 0.5      & $0.1518 \pm 0.0067$ & $0.1935 \pm 0.0192$ \\
    \midrule
    Ethnicity & baseline & $0.1188 \pm 0.0105$ & $0.1851 \pm 0.0128$ \\
              & 0.01     & $0.1206 \pm 0.0080$ & $0.1809 \pm 0.0276$ \\
              & 0.05     & $0.1199 \pm 0.0098$ & $0.1978 \pm 0.0184$ \\
              & 0.1      & $0.1196 \pm 0.0041$ & $0.1998 \pm 0.0122$ \\
              & 0.5      & $0.1209 \pm 0.0059$ & $0.2026 \pm 0.0213$ \\
    \bottomrule
  \end{tabular}
\end{table}

\FloatBarrier
\section{Fairness Metrics with Confidence Intervals}
\label{app:fairness-ci}

This appendix reports the full fairness results that support the primary
conclusions in Section~\ref{sec:results}. Two sets of numbers are
reported. The per-article tables below (Tables~\ref{tab:b1}--\ref{tab:b4gender}),
except the Disparate Impact summary (Table~\ref{tab:di}), are seed-42
estimates, in which all intervals are 95\% bootstrap confidence
intervals computed by resampling the 1{,}000 test documents with replacement
($n = 10{,}000$, seed 42) and taking the 2.5th and 97.5th percentiles. The
across-seed summary (Table~\ref{tab:b-multiseed}) and the Disparate Impact
table aggregate the five seeds and form the basis for the conclusions in the
main text. Per-article decision thresholds are tuned independently for each
model on the validation set, matching the main evaluation. Gender and ethnicity
protected groups are operationalised exactly as in Section~\ref{sec:eval}.

\begin{table}[H]
\centering
\caption{Across-seed fairness summary over the seven reliable articles
(mean over five seeds: 42, 13, 7, 21, 100). For each article, the table
reports the number of seeds (out of five) whose per-seed paired bootstrap
$\Delta$DPD interval crosses zero; a count near five indicates no detectable
change. Two ethnicity operationalisations are shown: keyword matching
(protected $n=136$) and named-entity recognition (NER; protected $n=59$;
Section~\ref{sec:discussion}).}
\label{tab:b-multiseed}

\footnotesize
\setlength{\tabcolsep}{2pt}
\renewcommand{\arraystretch}{1.1}

\begin{tabularx}{\columnwidth}{
    @{}
    l
    >{\centering\arraybackslash}X
    >{\centering\arraybackslash}X
    >{\centering\arraybackslash}X
    @{}
}
\toprule
Article &
\shortstack{Gender\\(keyword)\\cross/5} &
\shortstack{Ethnicity\\(keyword)\\cross/5} &
\shortstack{Ethnicity\\(NER)\\cross/5} \\
\midrule
Art.2  & 5 & 5 & 3 \\
Art.3  & 3 & 5 & 5 \\
Art.5  & 2 & 5 & 4 \\
Art.6  & 4 & 3 & 3 \\
Art.8  & 4 & 5 & 4 \\
Art.10 & 5 & 2 & 3 \\
P1-1   & 5 & 4 & 3 \\
\midrule
Mean cross/5 & 5.6 & 5.8 & 5.0 \\
\bottomrule
\end{tabularx}
\end{table}

\noindent
The per-seed crosses-zero counts underlying the means above are
$6,6,6,5,5$ (gender, keyword), $6,7,5,5,6$ (ethnicity, keyword), and
$5,6,3,6,5$ (ethnicity, NER). The NER grouping resolves ethnicity membership
through named-entity recognition rather than whole-word matching, yielding
a smaller protected group (59 versus 136 documents). It reproduces the
aggregate null, with no article excluding zero consistently across seeds
under either operationalisation.

\subsection{Change in DPD under regularisation (paired bootstrap, seed 42)}
\label{app:delta-dpd}

The central question for the fairness null is not whether each model is
unfair, but whether SCM regularisation changes the disparity.
Table~\ref{tab:b1} reports $\Delta$DPD (SCM minus baseline) per article with a
paired bootstrap confidence interval, in which the same resampled document
indices are applied to both models on each iteration. This pairing accounts
for the correlation between the two models' errors and isolates the effect of
the intervention. An interval that contains zero indicates that the observed
change is not distinguishable from sampling noise. The table reports the
seed-42 estimates; the across-seed pattern is summarised in
Table~\ref{tab:b-multiseed} and Section~\ref{sec:results}.

\begin{table}[H]
\centering
\caption{Change in DPD (SCM, $\lambda = 0.1$, minus baseline) per article,
seed 42, with 95\% paired bootstrap CI. Negative $\Delta$DPD indicates improved
parity. An interval containing 0 indicates no statistically detectable change.
$\dagger$ = unreliable ($\leq 37$ ground-truth positive test labels).}
\label{tab:b1}

\footnotesize
\setlength{\tabcolsep}{2.5pt}
\renewcommand{\arraystretch}{1.08}

\resizebox{\columnwidth}{!}{%
\begin{tabular}{lcccc}
\toprule
Article &
\shortstack{Gender\\$\Delta$DPD} &
\shortstack{Gender\\95\% CI} &
\shortstack{Ethnicity\\$\Delta$DPD} &
\shortstack{Ethnicity\\95\% CI} \\
\midrule
Art.2  & $-0.0032$ & $[-0.0087,+0.0075]$ & $-0.0050$ & $[-0.0231,+0.0070]$ \\
Art.3  & $-0.0140$ & $[-0.0257,-0.0022]$ & $+0.0150$ & $[-0.0174,+0.0461]$ \\
Art.5  & $-0.0247$ & $[-0.0719,+0.0097]$ & $-0.0325$ & $[-0.0821,+0.0163]$ \\
Art.6  & $-0.0231$ & $[-0.0998,+0.0497]$ & $-0.0590$ & $[-0.0994,-0.0190]$ \\
Art.8  & $+0.0086$ & $[-0.0032,+0.0203]$ & $+0.0178$ & $[-0.0272,+0.0510]$ \\
Art.10 & $-0.0107$ & $[-0.0182,+0.0149]$ & $+0.0054$ & $[-0.0135,+0.0301]$ \\
P1-1   & $+0.0043$ & $[-0.0361,+0.0460]$ & $+0.0039$ & $[-0.0104,+0.0227]$ \\
\midrule
Art.9$\dagger$  & $-0.0011$ & $[-0.0033,+0.0000]$ & $-0.0074$ & $[-0.0242,+0.0000]$ \\
Art.11$\dagger$ & $-0.0048$ & $[-0.0232,+0.0245]$ & $+0.0031$ & $[-0.0124,+0.0162]$ \\
Art.14$\dagger$ & $+0.0038$ & $[-0.0308,+0.0228]$ & $+0.0038$ & $[-0.0227,+0.0189]$ \\
\bottomrule
\end{tabular}%
}
\end{table}

In this seed, on the ethnicity axis, where the protected group is small but
large enough for stable per-article estimates, the change in DPD is not
statistically distinguishable from zero for six of the seven reliable
articles. The only interval excluding zero here is Art.6
$[-0.0994, -0.0190]$. Across all five seeds, however, Art.6 excludes zero in
only two seeds and with inconsistent magnitude, so it does not constitute a
reproducible effect (Table~\ref{tab:b-multiseed},
Section~\ref{sec:results}). The gender axis shows the same pattern
with far wider intervals, reflecting the 69-document unprotected group: only
Art.3 $[-0.0257, -0.0022]$ excludes zero, and larger nominal changes such as
Art.6 $[-0.0998, +0.0497]$ span zero. Taken together, the paired intervals
provide inferential support for the null reported in
Section~\ref{sec:results}.

\FloatBarrier
\subsection{DPD with confidence intervals (seed 42)}
\label{app:dpd-ci}

Table~\ref{tab:b2gender} and Table~\ref{tab:b2eth} report per-article DPD for
both models with 95\% bootstrap confidence intervals, seed 42. The unreliable
articles (Art.9, Art.11, Art.14) are included here, as their bootstrap
intervals are promised in Section~\ref{sec:eval}. On the gender axis, the
reliable-article intervals are noticeably wider than their ethnicity
counterparts, reflecting the 69-document unprotected group.

\begin{table}[H]
\centering
\caption{Gender DPD for baseline and SCM ($\lambda = 0.1$), seed 42, with 95\%
bootstrap CIs. $\dagger$ = unreliable.}
\label{tab:b2gender}
\small
\begin{tabular}{lcccc}
\toprule
Article & DPD base & 95\% CI & DPD SCM & 95\% CI \\
\midrule
Art.2  & 0.0145 & $[0.0010, 0.0610]$ & 0.0113 & $[0.0009, 0.0588]$ \\
Art.3  & 0.2191 & $[0.1671, 0.2639]$ & 0.2052 & $[0.1537, 0.2494]$ \\
Art.5  & 0.1520 & $[0.0930, 0.2014]$ & 0.1273 & $[0.0572, 0.1892]$ \\
Art.6  & 0.3176 & $[0.1989, 0.4300]$ & 0.2945 & $[0.1749, 0.4121]$ \\
Art.8  & 0.0838 & $[0.0327, 0.1220]$ & 0.0924 & $[0.0416, 0.1311]$ \\
Art.10 & 0.0354 & $[0.0018, 0.1092]$ & 0.0247 & $[0.0014, 0.0995]$ \\
P1-1   & 0.3247 & $[0.2052, 0.4411]$ & 0.3290 & $[0.2123, 0.4480]$ \\
\midrule
Art.9$\dagger$  & 0.0097 & $[0.0043, 0.0164]$ & 0.0086 & $[0.0032, 0.0151]$ \\
Art.11$\dagger$ & 0.0145 & $[0.0009, 0.0379]$ & 0.0097 & $[0.0009, 0.0470]$ \\
Art.14$\dagger$ & 0.0177 & $[0.0011, 0.0728]$ & 0.0215 & $[0.0025, 0.0672]$ \\
\bottomrule
\end{tabular}
\end{table}

\begin{table}[H]
\centering
\caption{Ethnicity DPD for baseline and SCM ($\lambda = 0.1$), seed 42, with
95\% bootstrap CIs. Notation as above.}
\label{tab:b2eth}
\small
\begin{tabular}{lcccc}
\toprule
Article & DPD base & 95\% CI & DPD SCM & 95\% CI \\
\midrule
Art.2  & 0.1127 & $[0.0525, 0.1771]$ & 0.1077 & $[0.0495, 0.1715]$ \\
Art.3  & 0.0792 & $[0.0073, 0.1620]$ & 0.0943 & $[0.0158, 0.1773]$ \\
Art.5  & 0.1944 & $[0.1105, 0.2771]$ & 0.1619 & $[0.0804, 0.2463]$ \\
Art.6  & 0.3188 & $[0.2576, 0.3766]$ & 0.2597 & $[0.1997, 0.3175]$ \\
Art.8  & 0.0217 & $[0.0012, 0.0708]$ & 0.0395 & $[0.0025, 0.0877]$ \\
Art.10 & 0.0652 & $[0.0147, 0.1219]$ & 0.0706 & $[0.0166, 0.1301]$ \\
P1-1   & 0.1106 & $[0.0631, 0.1527]$ & 0.1144 & $[0.0683, 0.1544]$ \\
\midrule
Art.9$\dagger$  & 0.0492 & $[0.0143, 0.0897]$ & 0.0418 & $[0.0101, 0.0797]$ \\
Art.11$\dagger$ & 0.0069 & $[0.0006, 0.0322]$ & 0.0099 & $[0.0007, 0.0391]$ \\
Art.14$\dagger$ & 0.0113 & $[0.0007, 0.0473]$ & 0.0151 & $[0.0006, 0.0424]$ \\
\bottomrule
\end{tabular}
\end{table}

\FloatBarrier
\subsection{Disparate Impact}
\label{app:di-ci}

Disparate Impact (the ratio of positive-prediction rates between
groups) is reported in Table~\ref{tab:di} as a secondary diagnostic
but is not used for any primary conclusion. On the gender axis the
small unprotected group (69 documents) makes the ratio unstable: the
across-seed standard deviation is large relative to the effect
($3.56 \pm 0.73$ under SCM), and per-article DI is undefined or
extreme whenever a group has few positive predictions. On the more
stable ethnicity axis DI is essentially unchanged under SCM
($1.54 \pm 0.10 \rightarrow 1.48 \pm 0.08$), consistent with the DPD
and EOD nulls. Because of this instability, we rely on DPD and EOD for
all fairness conclusions.

\begin{table}[h]
\caption{Disparate Impact (DI) over the reliable articles, mean $\pm$
std across five seeds. DI is the ratio of positive-prediction rates
(protected / unprotected); $\text{DI}=1$ indicates parity and the
four-fifths rule flags $\text{DI}<0.8$ as adverse impact. DI is reported
as a diagnostic only: on the gender axis the small unprotected group
(69 documents) makes the ratio unstable, reflected in the large
across-seed standard deviation, so no conclusion is drawn from it.}
\label{tab:di}
\begin{tabular}{lcc}
\toprule
Axis & DI baseline & DI SCM \\
\midrule
Gender    & $2.44 \pm 0.35$ & $3.56 \pm 0.73$ \\
Ethnicity & $1.54 \pm 0.10$ & $1.48 \pm 0.08$ \\
\bottomrule
\end{tabular}
\end{table}

\FloatBarrier
\subsection{Equalized Odds with confidence intervals (seed 42)}
\label{app:eod-ci}

EOD conditions on the ground-truth label and is reported here as the larger of
the per-group true-positive-rate gap and false-positive-rate gap. Higher
values indicate less equitable error distribution. On articles with few
protected positives the rate estimates rest on small samples, so the intervals
are wide. The tables report seed-42 estimates; the across-seed mean EOD per
model is reported in the main text (Table~\ref{tab:fairness_multiseed}).

\begin{table}[H]
\centering
\caption{Ethnicity EOD for baseline and SCM ($\lambda = 0.1$), seed 42, with
95\% bootstrap CIs.}
\label{tab:b4eth}
\small
\begin{tabular}{lcccc}
\toprule
Article & EOD base & 95\% CI & EOD SCM & 95\% CI \\
\midrule
Art.2  & 0.250 & $[0.111, 0.400]$ & 0.200 & $[0.034, 0.371]$ \\
Art.3  & 0.040 & $[0.013, 0.145]$ & 0.047 & $[0.017, 0.167]$ \\
Art.5  & 0.178 & $[0.085, 0.293]$ & 0.123 & $[0.045, 0.256]$ \\
Art.6  & 0.456 & $[0.158, 0.789]$ & 0.512 & $[0.194, 0.755]$ \\
Art.8  & 0.266 & $[0.043, 0.492]$ & 0.218 & $[0.036, 0.450]$ \\
Art.10 & 0.113 & $[0.014, 0.364]$ & 0.089 & $[0.015, 0.343]$ \\
P1-1   & 0.161 & $[0.088, 0.230]$ & 0.153 & $[0.078, 0.223]$ \\
\bottomrule
\end{tabular}
\end{table}

\begin{table}[H]
\centering
\caption{Gender EOD for baseline and SCM ($\lambda = 0.1$), seed 42, with 95\%
bootstrap CIs.}
\label{tab:b4gender}
\small
\begin{tabular}{lcccc}
\toprule
Article & EOD base & 95\% CI & EOD SCM & 95\% CI \\
\midrule
Art.2  & 0.170 & $[0.011, 0.278]$ & 0.189 & $[0.009, 0.297]$ \\
Art.3  & 0.117 & $[0.044, 0.159]$ & 0.160 & $[0.037, 0.208]$ \\
Art.5  & 0.226 & $[0.043, 0.290]$ & 0.238 & $[0.015, 0.303]$ \\
Art.6  & 0.201 & $[0.119, 0.289]$ & 0.180 & $[0.074, 0.294]$ \\
Art.8  & 0.103 & $[0.012, 0.658]$ & 0.112 & $[0.018, 0.664]$ \\
Art.10 & 0.589 & $[0.440, 0.700]$ & 0.521 & $[0.375, 0.634]$ \\
P1-1   & 0.210 & $[0.098, 0.345]$ & 0.216 & $[0.106, 0.351]$ \\
\bottomrule
\end{tabular}
\end{table}

Across both axes, the baseline and SCM EOD intervals overlap substantially for
every article, indicating that SCM regularisation does not produce a
detectable change in error-rate equity. This is consistent with the DPD null
and extends it to a label-conditioned fairness criterion. The fairness null
therefore holds across two distinct fairness definitions, not one.

\FloatBarrier
\section{SCM Antonym Pairs}
\label{app:scm-pairs}

Table~\ref{tab:scm-antonym-pairs} lists the antonym pairs used to construct
the SCM-based regularisation term. The pairs operationalise the two
Stereotype Content Model dimensions used in this paper: warmth and
competence. Of the 38 pairs (19 warmth, 19 competence), the core set follows
\citet{omrani2023scm}, extended with additional warmth and competence markers
drawn from \citet{Ayesh2026AnnotatingDO} to better cover the register of legal
text.

\begin{table}[H]
\centering
\caption{SCM antonym pairs used for the regularisation loss.}
\label{tab:scm-antonym-pairs}
\small
\begin{tabular}{lll}
\toprule
Dimension & Positive pole & Negative pole \\
\midrule
Warmth & sincere & dishonest \\
Warmth & trustworthy & untrustworthy \\
Warmth & kind & cruel \\
Warmth & warm & cold \\
Warmth & friendly & hostile \\
Warmth & generous & selfish \\
Warmth & caring & indifferent \\
Warmth & genuine & fake \\
Warmth & helpful & unhelpful \\
Warmth & tolerant & intolerant \\
Warmth & benevolent & malevolent \\
Warmth & compassionate & callous \\
Warmth & honest & deceptive \\
Warmth & loyal & disloyal \\
Warmth & trustful & suspicious \\
Warmth & sociable & antisocial \\
Warmth & cooperative & uncooperative \\
Warmth & respectful & disrespectful \\
Warmth & well-meaning & ill-intentioned \\
\midrule
Competence & intelligent & stupid \\
Competence & capable & incapable \\
Competence & skilled & unskilled \\
Competence & competent & incompetent \\
Competence & efficient & inefficient \\
Competence & qualified & unqualified \\
Competence & expert & amateur \\
Competence & proficient & inept \\
Competence & resourceful & helpless \\
Competence & able & unable \\
Competence & knowledgeable & ignorant \\
Competence & talented & talentless \\
Competence & experienced & inexperienced \\
Competence & reliable & unreliable \\
Competence & systematic & haphazard \\
Competence & methodical & chaotic \\
Competence & precise & imprecise \\
Competence & thorough & careless \\
Competence & coherent & incoherent \\
\bottomrule
\end{tabular}
\end{table}

\FloatBarrier
\section{Protected-Group Keyword Lists}
\label{app:keywords}

Protected groups are operationalised through whole-word regular-expression
matching of the term lists below against the document text, defined separately
for gender and ethnicity. A document is assigned to the protected group on an
axis if it contains at least one matching term on that axis. Under this
operationalisation the test set splits into 931 protected and 69 unprotected
documents for gender, and 136 protected and 864 unprotected documents for
ethnicity.

The ethnicity list is informed by the demographic lexicon of
\citet{choenni2021}, as adapted in \citet{kocadag2026assessing}, but is restricted to
ethnicity and minority-status terms and supplemented with groups salient in
ECtHR proceedings (Roma, Kurdish, Chechen, asylum-seeker, refugee). Nationality
adjectives present in the source lexicon (e.g.\ European, American, British)
were deliberately excluded: respondent states are named in nearly every ECtHR
judgment, so including them would assign roughly half the corpus to the
protected group and dissolve the minority comparison. The list is therefore
purpose-built for this corpus rather than adopted wholesale from a validated
source. An ethnicity NER variant (Section~\ref{sec:discussion}) resolves group
membership through named-entity recognition rather than whole-word
matching. Because NER requires a term to be recognised as an entity span
rather than a bare string match, it yields a smaller protected group
(59 versus 136 documents), and serves as a robustness check on the
keyword operationalisation.

\begin{table}[H]
\centering
\caption{Protected-group keyword lists. Gender: 40 terms; ethnicity: 23 terms.}
\label{tab:keywords}

\footnotesize
\setlength{\tabcolsep}{3pt}
\renewcommand{\arraystretch}{1.08}

\begin{tabularx}{\columnwidth}{
    @{}
    p{0.20\columnwidth}
    X
    @{}
}
\toprule
Axis & Terms \\
\midrule

Gender (40) &
woman, women, female, girl, mother, wife, daughter, sister, she, her, hers,
lady, bride, girlfriend, stepmother, grandmother, schoolgirl, mommy, aunt,
niece, man, men, male, boy, father, husband, son, brother, he, him, his,
gentleman, groom, boyfriend, stepfather, grandfather, schoolboy, daddy,
uncle, nephew \\

\midrule

Ethnicity (23) &
roma, romani, gypsy, kurdish, kurd, chechen, jewish, muslim, christian,
orthodox, asylum, refugee, immigrant, migrant, foreigner, minority, ethnic,
ethnicity, race, racial, indigenous, aboriginal, caste \\

\bottomrule
\end{tabularx}
\end{table}

\FloatBarrier
\section{Instance-Weighted Regularisation}
\label{app:iw}
A natural objection to the fairness null is that uniform
regularisation applies equal pressure to every document,
regardless of its stereotypical content. To test this, each
document's SCM loss was weighted by its share of detected
antonym pairs, so that the most stereotype-laden documents
received the strongest regularisation. The weights varied
substantially across the training set (mean 0.23, standard
deviation 0.13). Even so, instance weighting did not produce a
consistent fairness improvement. Over the seven reliable articles
the mean absolute DPD is essentially unchanged on both axes
(gender $+0.006$, ethnicity $-0.007$), and the per-article signs
are mixed. On ethnicity, two reliable articles show single-seed
point-estimate reductions beyond the pre-registered threshold
(Art.~5 $\Delta\mathrm{DPD} = -0.051$, Art.~6 $\Delta\mathrm{DPD} = -0.057$),
but these are uncorrected point estimates from a single seed
without bootstrap confidence intervals, and the mean over reliable
articles remains flat with no consistent direction across articles.
Performance was unaffected (macro F1 = 0.6758). The fairness null
therefore survives the most direct refinement of the intervention
at the aggregate level. This strengthens the interpretation that
disparity in ECtHR predictions is driven by factors beyond the
token-level stereotypical associations that SCM targets.

\FloatBarrier
\section{Cross-Encoder Replication}
\label{app:cross-encoder}

To test whether the faithfulness finding depends on LegalBERT specifically, the baseline-versus-SCM comparison was replicated on \texttt{bert-base-uncased} and \texttt{roberta-base}, three seeds each, through the identical pipeline. These runs check the direction of the sufficiency effect across encoder families rather than claiming uniform replication. The degradation appears in four of the six encoder-seed runs. It is strongest on RoBERTa, where mean sufficiency rises from $0.116 \pm 0.028$ to $0.159 \pm 0.060$ at $k = 5\%$ (mean increase $+0.043$, two of three seeds), comparable in size to the LegalBERT effect. On general-domain BERT the change is weak and mixed: the mean increase at $k = 5\%$ is near zero ($+0.0003$) and one of the three seeds improves rather than degrades. The degradation is therefore not specific to the legal encoder, present in the majority of runs and clearest on RoBERTa, while neither uniform across encoders nor consistent on BERT. The across-seed variance on both additional encoders is large given only three seeds, so these results are supporting evidence rather than a second primary result.

\begin{figure}[H]
  \centering
  \includegraphics[width=\linewidth]{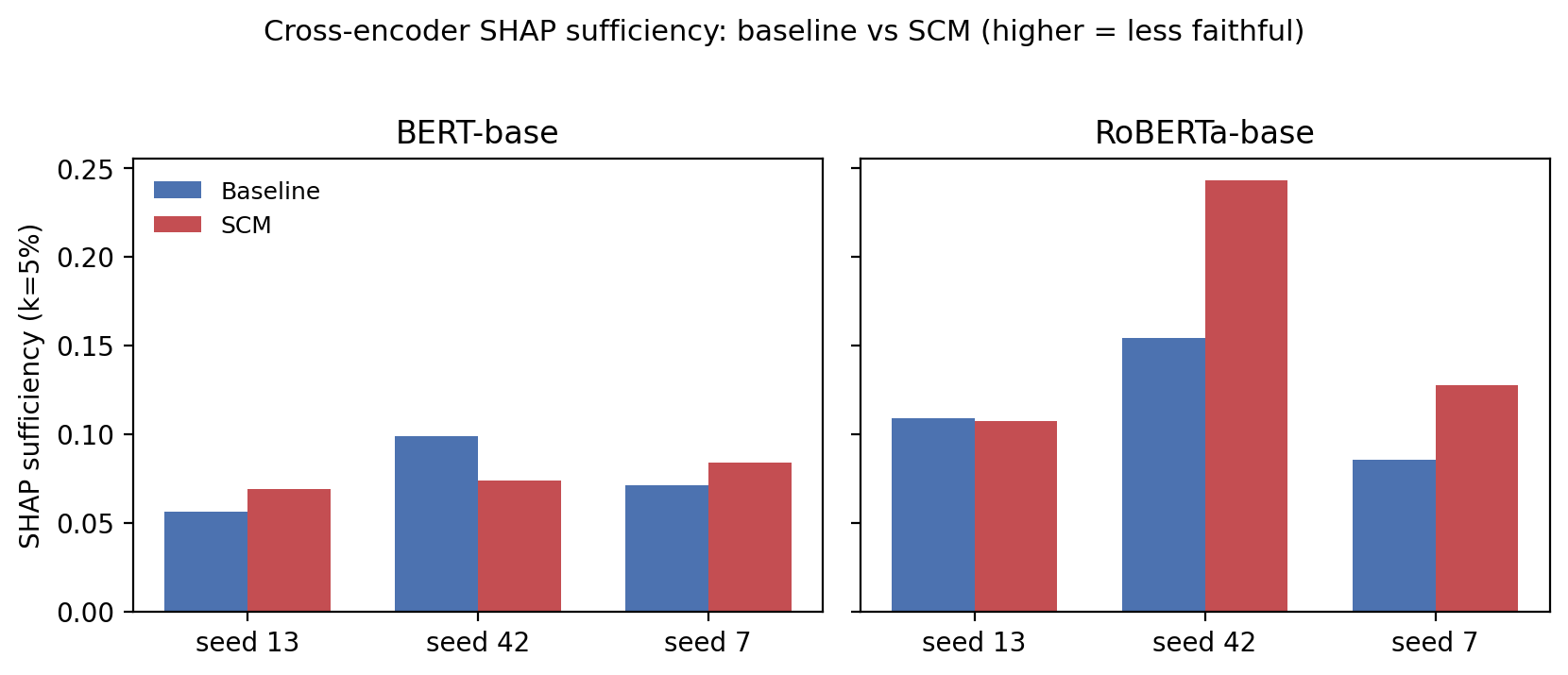}
  \caption{Cross-encoder SHAP sufficiency (baseline vs SCM) at $k=5\%$ on
  \texttt{bert-base-uncased} and \texttt{roberta-base}, three seeds each.
  Higher sufficiency indicates a less faithful explanation. The degradation
  is clearest on RoBERTa (mean $0.116 \to 0.159$) and weak and mixed on BERT,
  where one seed improves. Four of the six encoder--seed runs show the
  degradation.}
  \label{fig:apx:cross-encoder}
\end{figure}

\FloatBarrier
\section{Demographic Word-Pair Debiasing Control}
\label{app:wordpair}
\label{sec:wordpair-results}

The shuffled-pair control rules out the SCM antonym structure as the source of
the effects, but leaves open whether a conventional demographic word-pair
debiasing method~\cite{bolukbasi2016debias} would have moved disparity where the
SCM term did not. The pipeline was repeated with a demographic word-pair
condition using the identical contrastive loss, margin ($m = 0.5$), and
$\lambda$ grid as the SCM condition (Equations~\ref{eq:loss}--\ref{eq:scm}),
substituting demographic pairs for the warmth--competence antonyms. Each axis
was debiased separately and evaluated against its own freshly trained baseline
under the same five seeds. The gender axis uses nineteen genuine binary
opposites (\emph{man}--\emph{woman}, \emph{he}--\emph{she}, and so on), the
canonical target of word-pair debiasing. The ethnicity axis has no natural antonyms, so each of sixteen minority terms is contrasted with an unmarked majority-referent term (e.g.\ citizen, national, resident); this minority--majority construction is deliberately artificial, included to test whether the approach applies to the ethnicity axis at all. 

Table~\ref{tab:wordpair} reports mean absolute DPD over the seven reliable
articles per seed and axis. No seed or axis reaches the pre-registered
$\Delta\text{DPD}\ge 0.05$ threshold. On gender the per-seed changes are
$+0.008, -0.023, +0.004, -0.004, +0.008$ (mean $-0.001$) and on ethnicity
$+0.012, +0.009, -0.004, -0.001, -0.017$ (mean $-0.000$); on both axes the sign
is inconsistent across seeds. Classification performance is unaffected
(mean $\Delta\text{F1}=-0.002$). This adds a third mechanistically distinct
intervention, after SCM regularisation and the shuffled-pair control, that
leaves ECtHR disparity unchanged. That a genuine word-pair method also fails to
move DPD on the gender axis, where the paradigm is designed to apply, indicates
the fairness null is not an artefact of the SCM design or of contrastive
regularisation in particular.

\begin{figure}[H]
  \centering
  \includegraphics[width=\linewidth]{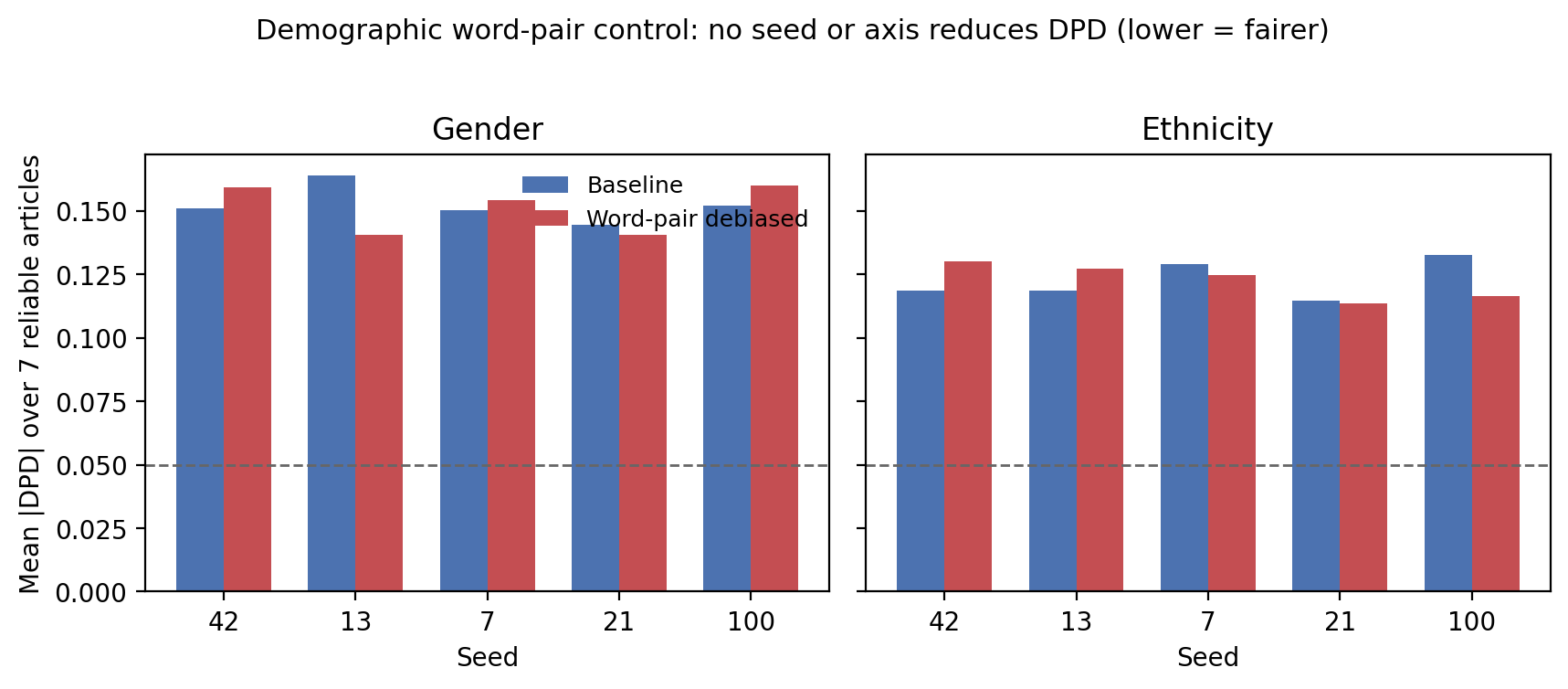}
  \caption{Demographic word-pair control: mean $|$DPD$|$ over the seven
  reliable articles, baseline versus word-pair-debiased, per seed and axis.
  The dashed line marks $\Delta\mathrm{DPD}=0.05$. No seed or axis reduces
  disparity below baseline by a practically meaningful margin; the companion
  per-seed values are in Table~\ref{tab:wordpair}.}
  \label{fig:apx:wordpair}
\end{figure}

\begin{table}[H]
\centering
\caption{Mean absolute DPD over the seven reliable articles for the baseline and
the demographic word-pair-debiased model, per seed, with the validation-selected
$\lambda$. No seed or axis reaches the pre-registered
$\Delta\text{DPD}\ge 0.05$ threshold.}
\label{tab:wordpair}
\begin{tabular}{llccc}
\toprule
Axis & Seed ($\lambda$) & Base & Debiased & $\Delta$ \\
\midrule
\multirow{6}{*}{Gender}
 & 42 ($0.10$)  & 0.15 & 0.16 & $+0.01$ \\
 & 13 ($0.05$)  & 0.16 & 0.14 & $-0.02$ \\
 & 7  ($0.50$)  & 0.15 & 0.15 & $+0.00$ \\
 & 21 ($0.10$)  & 0.14 & 0.14 & $-0.00$ \\
 & 100 ($0.50$) & 0.15 & 0.16 & $+0.01$ \\
 & Mean         & 0.15 & 0.15 & $-0.00$ \\
\midrule
\multirow{6}{*}{Ethnicity}
 & 42 ($0.05$)  & 0.12 & 0.13 & $+0.01$ \\
 & 13 ($0.01$)  & 0.12 & 0.13 & $+0.01$ \\
 & 7  ($0.01$)  & 0.13 & 0.12 & $-0.00$ \\
 & 21 ($0.50$)  & 0.11 & 0.11 & $-0.00$ \\
 & 100 ($0.01$) & 0.13 & 0.12 & $-0.02$ \\
 & Mean         & 0.12 & 0.12 & $-0.00$ \\
\bottomrule
\end{tabular}
\end{table}

\FloatBarrier

\clearpage

\end{document}